# Uncertainty Management for Fuzzy Decision Support Systems

*Christoph F. Eick*

*University of Houston*
*Department of Computer Science*
*TX 77004*
*cs-net: ceick@houston*

**Abstract:** *A new approach for uncertainty management for fuzzy, rule based decision support systems is proposed. The domain expert's knowledge is expressed by a set of rules that frequently refer to vague and uncertain propositions. The certainty of propositions is represented using intervals [a b] expressing that the proposition's probability is at least a and at most b. Methods and techniques for computing the overall certainty of fuzzy compound propositions that have been defined by using logical connectives 'and', 'or' and 'not' are introduced. Different inference schemas for applying fuzzy rules by using modus ponens are discussed. Different algorithms for combining evidence that has been received from different rules for the same proposition are provided. The relationship of the approach to other approaches is analyzed and its problems of knowledge acquisition and knowledge representation are discussed in some detail. The basic concepts of a rule-based programming language called PICASSO, for which the approach is a theoretical foundation, are outlined.*

**Keywords and Phrases:** *rule based programming, decision support systems, interval valued logic, approximate reasoning, modus ponens generating functions, combination of evidence.*

## 1. Introduction

Very often in human life we are forced to make *guesses* in order to decide where certain objects are located, for reconstructing events that happened in the past or for building plans in order to achieve a certain goal. The human capability of *making complex decisions* is one of the most fascinating facets of human intelligence, especially if vague and uncertain knowledge is involved. Usually, decision making involves multiple knowledge sources from which the expert extracts different *clues* by using a set of frequently fuzzy *rules*, which encode the expert's general and domain-specific knowledge. Afterwards, the expert comes up with a decision by combining the evidence received from the individual clues.

It seems attractive for computerizing human decision making to have programming paradigms available that correspond to the expert's rule based inference process. Consequently, the use of *rule based programming paradigms* has gained a high popularity in recent research projects, especially in the area of expert systems. In these approaches the expert's knowledge is represented in sets of condition/action pairs.

```
if <conditions> then <actions>
```

The focus of this paper is the discussion of models, techniques and mechanisms that are suitable to automate human decision making involving vague and uncertain knowledge assuming a rule based programming environment as the underlying knowledge representation framework. The paper has the following organization. Section 2 surveys our model for representing uncertain and vague knowledge in a rule based programming environment. The sections 3, 4 and 5 provide logical operators for computing certainty factors of compound fuzzy propositions. Section 6 discusses an example and section 7 summarizes the results of the paper. Furthermore, the relationship of our approach to other approaches and methods for uncertainty management will be outlined in the subsequent sections.

The uncertainty management model, discussed in the paper, is a theoretical foundation of a rule-based programming language called PICASSO([EF 87]), whose main objectives can be characterized as follows:
- To facilitate the implementation of production rule oriented problem solving methods.
- To support the combination/integration of rules-based and functional programming styles and is completely compatible with LISP.
- To provide powerful mechanisms for representing fuzzy knowledge and for automating reasoning involving vague, uncertain or incomplete knowledge.
- To provide an expressive data model for the definition of permanent knowledge bases and powerful mechanisms for enforcing the consistency of the knowledge base.
- To support pattern directed invocation schemas by incorporating match- and other variables to PICASSO.

# 2. The Uncertainty Management Model of PICASSO

Decision making usually involves uncertain and incomplete knowledge that comes from different knowledge sources. Furthermore, different rules used in the inference process may have different degrees of empirical confirmation. Consider, the following example involving two rules:

(R1) infers, depending on the presence of certain empirical observations, that decision D is correct with a probability of 0.8 — R1 is a very reliable rule confirmed by empirical tests.

(R2) infers, if certain empirical observations are present, that decision D is correct with a probability of 0.2 — R2 is a vague rule of thumb that is used when not much knowledge is available.

If the rules R1 and R2 are both applicable, we would still say that D is correct in – let's say – about 75% of the cases, because the rule R1 is much more reliable than R2: that is, the evidence provided by R1 should outscore the evidence received by R2. On the other hand, frequently the empirical observations that are needed for applying a rule(in our example R1) are not available or too expensive to obtain (consider for example medical tests for which the patient has to pay); in this case, we are forced to make a less reliable decision based on R2: however, this is still better than rolling the dices.

Nearly all existing commercial expert system shells (for survey see [WAT 85]) as well as extensions of PROLOG ([HIN 86],[MUK 87],[VEM 86],[SLOP]) use " single-valued approaches" for reasoning in uncertainty; that is, they assign a probability-like value to each predicate of interest. However, single-valued approaches are not suitable for decision support systems because it is not possible to express the reliability of rules and predicates[1]. Furthermore, single-valued approaches have problems to represent the fact that we know nothing about a predicate P; in most cases, they identify 'unknown' with the probability 0.5 — for example, MYCIN([BUC 84]) uses this approach — which causes severe problems of knowledge representation.

Therefore, we have chosen a two-valued approach for representing fuzzy predicates, which is called *interval approach* as the underlying model; the model has been propagated — in different versions — by several research institutions([APB 85],[WIE 85],[LU 84]) to overcome the disadvantages of the single valued approach. The basic idea of the interval approach is to use upper and lower probability bounds to approximate the certainty of propositions. The assumptions of the interval approach are compatible with Dempster/Shafer's theory of evidence([SHA 76]). Our version of the interval approach can be characterized as follows:

The *belief* that a certain proposition P is true is measured by assigning an interval [a b] to P, expressing the following semantics:

(1) The probability that P is true is at least a. The *confirmation* of P is a: $\text{conf}[a\ b] = a$

(2) The probability that P is false is at least $(1-b)$. The *disconfirmation* of P is $(1-b)$: $\text{disconf}([a\ b]) = 1-b$.

(3) The *uncertainty* of our belief concerning P is measured by $(b-a)$: $\text{unc}([a\ b]) = b-a$.

(4) The *certainty* of our belief concerning P is measured by $1 - (b-a)$: $\text{cert}([a\ b]) = 1 - \text{unc}([a\ b])$.

(5) The *mean value* of our belief concerning P is measured by $\frac{a+b}{2}$: $\text{mv}([a\ b]) = \frac{a+b}{2}$

For example, if we assign an interval [40 99] to P we express the following: P is confirmed with 40%. The disconfirmation of P is 1%. That is, 40% of the probability is assigned to P, 1% of the probability is assigned to (not P); it is unknown how the remaining probability(59%) is distributed: we don't know how much of this probability is assigned to P and how much is assigned to (not P); the uncertainty is 59%. A special case is the interval [0 1]; it expresses the fact that we know nothing about a proposition P[2]; the confirmation and the disconfirmation of P is 0%; the uncertainty is 100%.

However, intervals are not only used for describing fuzzy predicates, but they are also used for specifying fuzzy rules using the following syntax[3]:

```
(<rule-id> (if <left-hand-side>) (then (infer <predicate> with <interval>)))
```

for example:

```
(r1 (if (tall ?x)) (then (infer (strong ?x) with (0.5 0.9))))
```

The latter PICASSO rule expresses: *If it is true that someone is tall, then the probability that he is strong is at least 0.5 and at most 0.9*

---

[1] Coming back to the above example, the single valued approaches can only assign a probability to the two rules(0.8 and 0.2, respectively) but they cannot express different degrees of reliability for the stored probabilities.

[2] That is, "unknown" can directly be represented in the interval approach.

[3] Variables prefixed by '?' represent match-variables.

The left-hand side of a PICASSO-rule consists of predicates connected by or-, and- or not-operators. For example,

```
(r2 (if (and (not(tall ?x))(heavy ?x)))
(then (infer (strong ?x) with (0.3 0.9))))
```

expresses: *if someone is not tall and heavy, then the probability that he is strong is at least 0.3 and at most 0.9*. In contrast to the first rule, the second rule provides only mildly suggestive positive evidence.

Surprisingly, although the interval approach has gained some popularity, a comprehensive methodology for automating inference in a rule-based environment is missing far for this approach. The remainder of the paper intends to provide such a methodology.

The problem of uncertain inference in a rule-based environment can be subdivided into 3 subproblems, which will be covered in the next three sections:

a) We need algorithms to compute the certainty factor of the left hand side of a rule, which usually consists of a set of conditions connected by and-, or- and not-operations.

b) Decisions criteria that decide when the right hand side of a rule is evaluated, and functions that compute the certainty factors of the inferred predicates have to be provided.

c) Combination algorithms have to be offered that combine the evidence inferred by different rules for the same predicate.

## 3. Computing the Certainty Factors of the Left-hand Side

In order to compute the uncertainty interval I for the left hand side of a rule, we have to be capable to apply logical operators 'and', 'or' and 'not' to intervals $I_1, ..., I_n$ measuring the uncertainty of each condition.

In order to compute the certainty factor of $A \wedge B$ correctly, we have to know something about the relationship of A and B; especially if A and B are positively correlated, uncorrelated or negatively correlated; otherwise, the estimation for $A \wedge B$ may lead to significant errors or very uncertain results. Surprisingly, most expert systems and shells evaluate the left-hand-side using the and/or-operators of Fuzzy Logic([KAN 86],[ZAD 85]), which assume that A and B maximally positively correlate(correlation=1). Although this assumption leads to nice theoretical properties (for example, the distributive law still holds in this case), it likely results in large calculation errors, if the predicates of the left hand side are uncorrelated or even negatively correlated[4].

Another research project([APB 85]) uses the interval approach but does not make any assumptions concerning the correlation of A and B; that is, the lower and upper bound of the resulting interval is calculated separately, choosing the correlation so that the lower bound becomes minimal and the upper bound becomes maximal. Because the approach does not use any knowledge about the correlation of A and B very uncertain intervals are received, when using this approach. Furthermore, frequently the approach computes the lower bound and the upper bound of an interval assuming different correlations in the two calculations, which brings up doubts about the overall consistency of the approach.

On the other hand, frequently the domain expert has a good feeling for the correlation of predicates used in a rule. Therefore, we have chosen approach that allows the PICASSO-programmer to assign a correlation to a rule. If such knowledge is not available the system will evaluate the left hand side of rules using a default correlation(correlation=0).

We distinguish the following cases for computing $P(A_1 ... \wedge ... \wedge A_n)$ assuming the certainty of $A_i$ is $[a_i \ b_i]$ for i=1,n.

case1: correlation=1 (best case)

$$A_1 \ [a_1 \ b_1]$$
$$...$$
$$A_n [a_n \ b_n]$$
$$\overline{A_1 \wedge ... \wedge A_n \ [min(a_1, ..., a_n), min(b_1, ..., b_n)]}$$

case2: correlation=0 – statistical independence

$$A_1 \ [a_1 b_1]$$
$$...$$
$$A_n [a_n b_n]$$
$$\overline{A_1 \wedge ... \wedge A_n \ [a_1 * ... * a_n, b_1 * ... * b_n]}$$

---

[4] Furthermore, this approach interprets the extreme case as the normal case: If we look to all predicates that exist in the cosmos — why should their average correlation be 1 and not something close to 0?

**case3:** correlation=-1 (worst case)

Let $\psi(c_1, ..., c_n) = max(0, 1 - (1 - c_1) - ... - (1 - c_n))$

$$A_1 \; [a_1 b_1]$$
$$A_1 \; [a_1 \; b_1]$$
$$...$$
$$A_n \; [a_n \; b_n]$$

$$\overline{A_1 \wedge ... \wedge A_n \;\; [\psi(a_1, ..., a_n), \psi(b_1, ..., b_n)]}$$

If the correlation has a value different from 1,0 or -1, we interpolate using the formulas of case1 and case2 or of case2 and case3, respectively.

For example, assuming correlation=0 we will receive for:

$(and \; (0.4 \; 0.9)(0.8 \; 0.9)) = (0.32 \; 0.81)$

Assuming correlation=1, we would receive:

$(and \; (0.4 \; 0.9)(0.8 \; 0.9)) = (0.4 \; 0.9)$

The negation of intervals can be computed using:

$$not(I) = not((a \; b)) = (1 - b \;\; 1 - a)$$

Finally, the corresponding formulas for the disjunction can easily be derived using the formulas for the conjunction and:

$$P(A \vee B) = P(A) + P(B) - P(A \wedge B)$$

Another special problem arises when the left-hand side of a rule refers to a predicate P that are not stored in the knowledge base. Two different interpretations make sense in this case:

1) *closed world assumption*: P is assumed to be false([0 0]).
2) *open world assumption*: P is assumed to be undefined([0 1]).

In contrast to most rule based programming languages, PICASSO uses the second, the open world assumption because of the following reasons:

- Decision making systems are frequently used in environments in which the knowledge evolves with time. That is, in the beginning of inference process many predicates are unknown. When using the closed world assumption these unknown predicates have be stored in the knowledge base, because otherwise they would be interpreted as false, which causes a significant overhead for the inference process, which does not occur when using the open world assumption.
- As a consequence of the last point, all predicates have to be known in advance[5], when using the closed world assumption, which is not the case, when using the open world assumption.

## 4. PICASSO's Modus Ponens Generating Functions

The next problem is, how can the positive/negative evidence a rule R1

(R1 (if E) (then (infer H with [c d])))

provides concerning a predicate H be computed, if the certainty factor of the rule's left-hand side has already been computed (using the methods described in section 3) as [a b]? Schematically:

$$\text{if E then H } [c \; d]$$
$$E \; [a \; b]$$

$$\overline{\textit{What is the rule's contribution } (?1 \; ?2) \textit{ concerning H?}}$$

Functions that compute (?1 ?2) are called *modus ponens generating functions*. The rest of this section will discuss different modus ponens generating functions that are compatible with the interval approach.

---

[5] Therefore, "closed world".

As mentioned at the beginning of the paper, the above PICASSO rules expresses: *"If E is known to be true, then the probability of H is at least c and at most d."* However, this interpretation doesn't tell us what to do in the case that E is false or only partially true. Two different extensions of the first interpretation are possible to deal with fuzzy predicates on the left-hand side of PICASSO-rules:

1) $P(H|E)$ is at least c and at most d
2) $P(E \longrightarrow H)$ is at least c and at most d

Depending on which of the two interpretations are chosen, different modus ponens generating functions can be derived. We will discuss each case separately.

## 4.1 Interpretation as a Conditional Probability

**assumption:** if E then H [c d] expresses: $P(H|E)$ *is at least c and at most d.*

Using this interpretation, we can derive the following modus ponens generating function[6]:

**Using**
*if E then H [c d]*
*E [a b]*
$\frac{a+b}{2} > \theta \wedge (b-a) < \psi$
**we infer:**
*H [?1 ?2]*
With

$$?1 = min(c * a + (1-d) * (1-a), c * b + (1-d) * (1-b))$$

$$?2 = min(1, max(d * a + (1-c) * (1-a), d * b + (1-c) * (1-b)))$$

**reasons:** Let E' denote the observations that cause to suspect that E is true and let $\bar{E}$ denote the a posteriori probability of E inferred taking into account the additional observations E'. Then, the above formula can be derived as follows:

$P(H|E') = P(H \wedge E|E') + P(H \wedge not(E)|E') =$

Making the "reasonable" assumption that "if we know that E is present(absent), then the observations E' relevant to E provide no further information about H" – that is, $P(H|E \wedge E') = P(H|E)$ and $P(H|E \wedge not(E)) = P(H|(not(E)))$ – we receive:

$$= P(E|E') * P(H|E' \wedge E) + P(not(E)|E') * P(H|not(E) \wedge E') =$$

$$P(H|E) * P(\bar{E}) + P(H|not(E)) * P(not(\bar{E}))$$

Unfortunately, $P(H|not(E))$ is unknown — we only can say that $P(H|E) > P(H|not(E))$ if E and H are positively correlated and $P(H|E) < P(H|not(E))$ if E and H are negatively correlated. Having no other clues concerning $P(H|not(E))$, we pragmatically set $P(H|not(E))$ to $P(H|E)$. The magnitude of the error we get by doing this is dependent on $(1-P(\bar{E}))$ and on $|correlation(E,H)|$. In order to keep the error small, rules will only be applied, if there is some positive evidence that E is true (for example we set $\theta = 0.55$ and $\psi = 0.85$).

Additionally taking into account that $P(\bar{E})$ is given as an interval [a b], $I_{\bar{E}}$ has to be computed so that the lower bound becomes minimal and the upper bound becomes maximal (varying possible values x with $a \leq x \leq b$ of $I_{\bar{E}}$). Taking into consideration that $c \leq P(H|E) \leq d$ holds, we receive for H's (a posteriori interval) [?1 ?2], described above.

An alternative approach used by the PROSPECTOR expert system([DUD 79]) requires that the expert also provides knowledge concerning $P(H|not(E))$. Obviously, in this case we would not get any error at all when using the formula provided before. Because the approach assumes that both relationships are given, PROSPECTOR rules are fired no matter what the probability of $P(\bar{E})$ is – in contrast to PICASSO's approach.

However, the approach has the severe disadvantage that it forces the expert to specify $P(H|not(E))$ although he might only have exact knowledge about $P(H|E)$. Let us illustrate our point giving a small common sense example: Obviously, $P(street\_wet|rain)=1$ because rain makes streets wet, but you may have severe problems to give an estimation for $P(street\_wet|not(rain))$.

---

[6] A similar modus ponens generating function has been proposed in [GIN 85].

Using PICASSO's approach a second rule r2, (r2 (if (not E)) then ((infer H ...)))) would be specified if knowledge is available concerning the relationship of not(E) and H.

Furthermore, MYCIN's modus ponens generating function is a special case of the above function. A MYCIN certainty factor mcf corresponds to an interval $[\frac{mcf+1}{2} \ 1]$ if mcf$\geq$ 0 and to an interval $[0 \ \frac{mcf+1}{2}]$ if mcf$\leq$ 0. By restricting the inputs to "MYCIN-intervals" ([0 a] or [b 1], where a and b are arbitrary real-numbers between 0 and 1), the above function degenerates to MYCIN's modus ponens generating function.

## 4.2 Interpretation as a Probability of a Disjunction

**general idea**: In contrast to the interpretation given before, this approach interprets *if E then H [c d]* as: *The probability of* $P(E \longrightarrow H)$ *is at least c and at most d.*

That is, this approach assigns an interval to:

$$P(A \longrightarrow B) = P(\neg A \vee B) = 1 - P(A) + P(B) - P(\neg A \wedge B)$$

Generally, the use of Modus Ponens may lead to inconsistencies — especially if mv(A) is close to 0. Again, distinguishing at least 3 cases depending on how the conjunction $\neg A \wedge B$ in the above formula is computed, we derive the following functions[7].

**case1**: uses Fuzzy Logic's and/or-function[8]:
**Using:**
If E then H [c d]
E [a b]
$(1 - a) \leq d$
**infer:**
if $(1 - a) < c$
 then H [c d]
 else H [0 d]

**case2**: assumes conventional probability theory is used[9]:
**Using:**
if E then H [c d]
E [a b]
$(a \neq 0 \wedge (c + a \geq 1) \wedge (b + d \geq 1))$
**infer:**
H $[\frac{c+a-1}{a} \ \frac{b+d-1}{b}]$

**case 3**: Use Fuzzy Logic's Implication-function, which assumes maximal disjointness[10]:
**Using:**
if E then H [c d]
E [a b]
$(a + c) \geq 1$
**infer:**
H [c+a-1 b+d-1]

The proofs of the 3 modus ponens generating functions are given in the long version of this paper.

The above 3 modus ponens generating functions are generalizations of functions published in [TRI 85], which were derived assuming a single valued approach. However, because of the incapability of single-valued approaches – already discussed in section 2 – to express the strength of belief in a proposition and the incability to represent "unknown" the published functions are not suitable for fuzzy decision making. For example, in the first modus ponens generating function if $c \geq (1 - a)$ holds, nothing can be inferred about the lower bound of $P(H)$; it is impossible to express this proposition in a single valued approach.

---

[7] A more detailed discussion and the proofs of the 3 modus ponens generating functions are given in the long version of the paper.

[8] Correlation=1 – the negated left hand side and the right hand side are are assumed to be maximally positively correlated.

[9] Correlation=0

[10] Correlation=-1 – it can easily be shown that in this case $P(A \longrightarrow B)$ is equal to max(1,1-P(A)+P(B)) – Fuzzy Logic's Implication-function(for definitions see [KAN 86]).

# 5. Combining Evidence Received from Several Rules

Several rules can provide positive or negative evidence concerning a predicate P. Assuming that n intervals $I_1, ..., I_n$ have been inferred using the methods discussed in sections 3 and 4, the overall certainty factor of P has to be computed by combining the evidence expressed in the intervals $I_1, ..., I_n$. Schematically:

$$R_1 \ provided \ evidence \ I_1 = (a_1 \ b_1) \ for \ predicate \ P$$
$$.............$$
$$R_n \ provided \ evidence \ I_n = (a_n \ b_n) \ for \ predicate \ P$$
$$Overall \ evidence \ I = (?1 \ ?2) \ for \ P$$

That is, we are looking for functions *combine* that give a reasonable estimation of the certainty I of P:

$$I = combine(I_1, combine(I_2, ..., combine(I_{n-1}, I_n)...))$$

## 5.1 Requirements for Good Combination Functions

Unfortunately, there is an unlimited number of possible combination functions *combine*. This brings up the question, which combination function is the best one. There is no unique answer to this question — reasoning is performed in different contexts. However, it is worth while to define some general requirements "good" candidate functions should meet:

(1) More certain clues should get a higher weight than clues that are less confirmed.
For example, the mean value of $combine([0.5 \ 0.9], [0.40 \ 0.41])$ should be relatively close to 0.41, because the second interval is 40 times less uncertain than the first one.

(2) The addition of very uncertain new evidence should not or only slightly affect the overall evidence concerning a predicate P. Especially, $combine([a \ b], [0 \ 1]) = [a \ b]$ must hold.

(3) A higher number of clues (with a similar uncertainty) should lead to a higher certainty of our estimates. In other words, the more evidence we have concerning a predicate of interest P, the more certain is our prediction. Furthermore, new evidence should never increase the uncertainty:
$cert(combine([a \ b], [c \ d])) \geq max\{cert([a \ b]), cert([c \ d])\}$

(4) The order in which certain evidence is combined should have no effect on the overall result; that is, if we combine evidence in different orders, then the result should be the same:

$combine(combine([a \ b], [c \ d]), [e \ f])) = combine(combine([e \ f], [a \ b]), [c \ d]))$

Remark: If this assumption is violated combining n intervals in different orders may lead to ambiguities, which is not desirable. Furthermore, if the above condition holds, it is sufficient to store for each proposition the current interval and not the set of previous intervals, which simplifies the implementation of the inference process a lot. On the other hand, this requirement puts severe restrictions on possible combination functions, because combination functions satisfying this requirement have to be associative and commutative.

(5) If we have received absolute certainty concerning a proposition, further new evidences should not change our overall perception of the proposition; that is we can stop making further inferences concerning P. Formally:
If $b \neq c$ then $combine([a \ a], [b \ c]) = [a \ a]$

Another important aspect – ignored so far in our discussions – is if the evidence concerning a predicate P has been received by interpreting the same or different knowledge sources.

Let's consider the following two examples: Two medical test T1 and T2 are used to predict the probability of P. Both tests interpret the same knowledge source (for example an ECG). T1 comes up with [70 80] and T2 comes up with [80 90]. In this case, we would expect the probability of P about 80% (we apply some kind of voting mechanism), because both intervals have been received from the same knowledge source. Furthermore, if T1 and T2 come up with more or less the same mean value, we would increase the strength of our belief in this probability.

On the other hand, let's assume the same intervals have been received concerning a predicate P by interpreting blood cultures(T1) and the patient's ECG(T2); that is, by interpreting two different knowledge

104

sources: In this case, we tend to give a much higher estimation for the probability of P(let's say 98%), because P has been confirmed by interpreting different (and not only single) knowledge sources.

Obviously, we have to distinguish these two cases in our requirements.

(6a) If rules interpreting different knowledge sources independently provide confirming positive(negative) and reliable evidence, then the mean value of the combined interval should be significantly greater(less) than the maximum mean value of the intervals to be combined: for example,

$$mv(combine([0.9\ 1], [0.8\ 1])) > 0.95$$
$$mv(combine([0\ 0.1], [0\ 0.2])) < 0.05$$

(6b) If evidence has been received using different rules interpreting the same knowledge source, then the combined mean value should not be significantly greater or less than the weighted average of the mean value of the intervals to be combined — using the uncertainty of the intervals as a weight.

From the discussion, we can draw the conclusion that we need at least two combination functions; one for combining evidence received from different knowledge sources and one for combining evidence that has been obtained by interpreting the same knowledge source.

Finding combination functions that satisfy all the requirements seems to be like solving a puzzle in which at least one piece is missing: Functions that satisfy 4 or 5 of the 6 requirements can easily be found; however, satisfying the last missing requirement(s) usually results in destroying other, already satisfied requirements.

The next subsection will introduce two combination functions that meet the discussed requirements to a large extend for the two different cases.

## 5.2 Two Combination Functions and their Properties

Two different combination functions are used in PICASSO:

- A function mscomb that assumes that the evidence has been received independently from different knowledge sources. That is, because of the independence assumption mscomb will increase the overall probability of P, if confirming evidence has been found. mscomb can be derived straight forward by applying Dempster's rule of combination[11].
- A new, so far unpublished, function sscomb that assumes that the evidence to be combined has been received by interpreting the same knowledge source; that is, the different intervals represent different "opinions" about the interpretation of the same data. Therefore, sscomb uses a weighted voting approach, that weights each interval to be combined by its uncertainty: certain intervals get a higher weight compared to uncertain intervals and confirming evidence will not increase the overall probability. For example, see the difference in:
sscomb([0.9 1],[0.9 1])=(0.925 0.975) and mscomb([0.9 1],[0.9 1])=[0.99 1].

mscomb and sscomb are defined as follows:

$$mscomb([l_1\ u_1], [l_2\ u_2]) =$$

$$[\frac{l_1 \times u_2 + l_2 \times u_1 - l_1 \times l_2}{1 - l_1 \times (1 - u_2) - l_2 \times (1 - u_1)} \quad \frac{u_1 \times u_2}{1 - l_1 \times (1 - u_2) - l_2 \times (1 - u_1)}]$$

Let $\tau = unc(I_1) + unc(I_2)$
Then sscomb is defined by the following two equations:

$$mv(sscomb(I_1, I_2)) = \begin{cases} 0.5 & \tau = 2 \\ \frac{cert(I_1)}{cert(I_1)+cert(I_2)} * mv(I_1) + \frac{cert(I_2)}{cert(I_1)+cert(I_2)} * mv(I_2) & 1 < \tau < 2 \\ \frac{unc(I_2)}{unc(I_1)+unc(I_2)} * mv(I_1) + \frac{unc(I_1)}{unc(I_1)+unc(I_2)} * mv(I_2) & 0 < \tau \leq 1 \\ \frac{mv(I_1)+mv(I_2)}{2} & \tau = 0 \end{cases}$$

---

[11] An interval [a b] approximating the certainty of a predicate P can be corresponds to the following probability assignment function: m($\emptyset$)=0, m(A)=a, m($\neg$A)= 1-b and m($\Omega$)=b-a. By applying Dempster's rule of combination mscomb can be derived immediately.

$$unc(sscomb(I_1, I_2)) = \begin{cases} 0 & if \ \tau = 0 \\ \frac{unc(I_1)*unc(I_2)}{unc(I_1)+unc(I_2)} & 0 < \tau \leq 1 \\ unc(I_1)*unc(I_2) & 1 < \tau < 2 \\ 1 & \tau = 2 \end{cases}$$

Obviously, the lower and upper bound of the combined interval I can easily computed using the mean value and the uncertainty.

The two combination functions have the following properties:

- In both functions the uncertainty of the combined evidence is smaller than the uncertainty of the original intervals: *many clues increase the certainty.*
- mscomb([0 1]),[a b])= sscomb([0 1],[a b])=[a b]: *completely uncertain knowledge should not change the overall evidence.*
- mscomb is associative; sscomb is associative when combining intervals, whose uncertainty is less equal 0.5.
- sscomb combines the intervals assigning a higher weight to more certain intervals; that is, it gives preference to evidence received from rules that have a high degree of confirmation. mscomb does not always have this property[12].
- MYCIN's combination function is a special case of mscomb.
- mscomb and sscomb are commutative.
  In summary, dscomb satisfies the requirements (2),(3),(4) and (6a); sscomb satisfies the requirements (1),(2),(3),(4) when combining certain intervals($\tau \leq 1$), (5) and (6b).

In [HUM 87] a combination function for combining dependent evidence has been proposed. However – in contrast to sscomb – this function assumes that the evidence to be combined has the same degree of confirmation. Because of this defect it doesn't seem very attractive to use this combination function in fuzzy decision support systems.

Another interesting aspect of using the interval approach is its capability to support non-monotonic reasoning processes. The classical Penguin-Bird-Fly example can be represented as:

```
(r1 (if (bird ?x)) (then (infer (fly ?x) with (0.999 1))))
(r2 (if (penguin ?y)) (then (infer (fly ?y) with (0 0))))
```

No matter which combination function we use, we receive the intended semantics: r2 overrules r1 in the case that both rules are fired.

## 6. A Small Example

We will finish this paper by giving a small example that demonstrates how the different functions discussed before work in practice: Consider we have the following PICASSO rule set:

```
(r1 (if (and (cloudy-sky)(humid))) (then (infer (rain) with (0.4 0.9))) (corr 0.5))
(r2 (if (high-pressure)) (then (infer (rain) with (0.0 0.7))))
(r3 (if (and (hot)(humid))) (then (infer (rain) with (0.6 1.0))) (corr 1.0))
(r4 (if (not(high-pressure)) (then (infer (rain) with (0.3 1.0))))
```

We assume the PICASSO-inference engine is called for the following 3 test-cases given below. We assume the inference engine's computation uses the following functions:

- mscomb is used for combining evidence.
- the modus ponens generating function of section 4.1 is used with $\theta$=0.55 and $\psi$=0.85.
- (hot) and (humid) have correlation 1; (cloudy-sky) and (humid) have 0.5.

---

[12] see for example:

dscomb([0.2 0.3],[0.0 0.8])=[0.166 0.250]

dscomb([0.2 0.3],[0.2 0.6])=[0.179 0.231]

dscomb([0.2 0.3],[0.4 0.4])=[0.200 0.200]

| Inferring (rain) | case1 | case2 | case3 |
|---|---|---|---|
| cloudy-sky | (0.88 0.90) | (0.60 0.62) | (0.90 0.92) |
| humid | (0.88 0.90) | (0.58 0.60) | (0.62 0.64) |
| hot | (0.80 0.82) | (0.90 0.94) | (0.65 0.67) |
| high-pressure | (0.32 0.34) | (0.80 0.82) | (0.90 0.92) |
| evidence r1 (rain) | (0.35 0.86) | not fired | (0.28 0.78) |
| evidence r2 (rain) | not fired | (0.05 0.76) | (0.02 0.73) |
| evidence r3 (rain) | (0.48 0.89) | (0.35 0.76) | (0.37 0.78) |
| evidence r4 (rain) | (0.20 0.90) | not fired | not fired |
| overall evidence (rain) | (0.67 0.84) | (0.32 0.64) | (0.40 0.60) |

We can interpret the results, after running the inference engine for the 3 test cases as follows:

- In the first case there is a strong suggestion that it will rain(about a 75% chance); in the other two cases a about 50% percent of rain chance was predicted. However, for the second test case this prediction is relatively unreliable: the uncertainty is 0.32.
- Note that in case 3 the rule r2 that provided positive evidence that balanced the negative evidence provided by rule r1 and r3.
- Although, the evidence we got in case 1 from r1, r2 and r4 was relatively uncertain, the presence of 3 independent clues supporting the rain hypothesis resulted in a relatively reliable predication of a 75% chance of rain.

# 7. Summary and Conclusion

The paper has focussed on the automation of reasoning processes involving vague and uncertain knowledge. It has demonstrated that the rules involved to solve typical decision making problems usually have different degrees of (empirical) confirmation, which are of critical importance for making the correct decision. This implies that at least a *two-valued* approach is required for automating fuzzy decision making: one value representing the probability p of a proposition P and the second value representing the strength of our belief that P has the probability p. An interval model for reasoning in uncertainty that meets these requirements has been provided. Using this model as an underlying framework,

- a correlation oriented, comprehensive approach for reasoning in uncertainty in a rule-based environment has been proposed.
- functions for computing the certainty factors involving logical connectives 'and', 'or' and 'not' and 4 different and new modus ponens generating functions have been provided.
- requirements for good combination functions were discussed and two combination functions, a new function for combining evidence involving the same knowledge source and a function for combining the evidence from different knowledge sources were introduced and their properties were discussed.

The methods and techniques discussed in the paper form the theoretical foundation for a rule-based programming language called PICASSO([EF 87]) and for a knowledge base management system called DALI([EICK 87]), which are currently under development at the University of Houston involving a group of 8 students.

Current and future research tries to test the suitability of our results by applying our methods to typical decision making problems found in card games – especially Bridge, software integration, computerized diagnosis and image interpretation.